\documentclass[10pt,twocolumn,letterpaper]{article}

\usepackage[
  pass,
]{geometry}

\usepackage{cvpr}              
\usepackage{diagbox}
\usepackage{graphicx}
\graphicspath{{FIGURES/}}
\usepackage[accsupp]{axessibility}
\usepackage[dvipsnames]{xcolor}

\definecolor{cvprblue}{rgb}{0.21,0.49,0.74}
\usepackage[pagebackref,breaklinks,colorlinks,citecolor=cvprblue]{hyperref}

\def\paperID{2832} 
\def\confName{CVPR}
\def\confYear{2024}

\title{Infrared Adversarial Car Stickers}

\author{Xiaopei Zhu$^{1}$ \ \ Yuqiu Liu$^{2}$ \ \ Zhanhao Hu$^{3}$ \ \  Jianmin Li$^{4}$ \ \ Xiaolin Hu$^{4,5,6}$\thanks{Corresponding author.}\\
$^{1}$School of Integrated Circuits, Tsinghua University, Beijing, China \\
$^{2}$Department of Technology, Beijing Forestry University, Beijing, China \\
$^{3}$Department of Electrical Engineering and Computer Sciences, UC Berkeley, California, USA \\
$^{4}$Department of Computer Science and Technology, Institute for Artificial Intelligence, \\ BNRist, Tsinghua University, Beijing, China \\
$^{5}$THBI, IDG/McGovern Institute for Brain Research, Tsinghua University, Beijing, China \\
$^{6}$Chinese Institute for Brain Research (CIBR), Beijing, China \\
\tt\small \{zxp18\}@mails.tsinghua.edu.cn \\ \tt\small \{liuyuqiu99, zhanhaohu.cs\}@gmail.com \\ \tt\small \{lijianmin, xlhu\}@mail.tsinghua.edu.cn}

\begin{document}
\maketitle
\begin{abstract}
Infrared physical 
adversarial examples are of great significance for studying the security of infrared AI systems 
that are widely used in our lives such as autonomous driving. Previous infrared physical attacks 
mainly focused on 2D infrared pedestrian detection which may not fully manifest its destructiveness to AI systems.
In this work, we propose a physical attack method against infrared detectors based 
on 3D modeling, which is applied to a real car.  
The goal is to design a set of infrared adversarial stickers to make cars invisible to infrared detectors at 
various viewing angles, distances, and scenes.
We build a 3D infrared car model with real infrared characteristics and 
propose an infrared adversarial pattern generation method based on 3D mesh shadow.
We propose a 3D control points-based mesh smoothing algorithm and use a set of smoothness loss 
functions to enhance the smoothness of adversarial meshes and facilitate the sticker implementation.
Besides, We designed the aluminum stickers 
and conducted physical experiments on two real Mercedes-Benz A200L 
cars. Our adversarial stickers hid the cars from Faster RCNN, an object detector, at various viewing angles, 
distances, and scenes. The attack success rate (ASR) was 91.49\% for real cars. 
In comparison, the ASRs of 
random stickers and no sticker were only 6.21\% and 0.66\%, respectively. 
In addition, the ASRs of the designed stickers against six unseen object detectors such as YOLOv3 
and Deformable DETR were between 73.35\%-95.80\%, showing good transferability of the attack performance across detectors. 

\end{abstract}

\section{Introduction}
\label{sec:intro}

\begin{figure}[tbp]
\centering
\includegraphics[width=1\columnwidth]{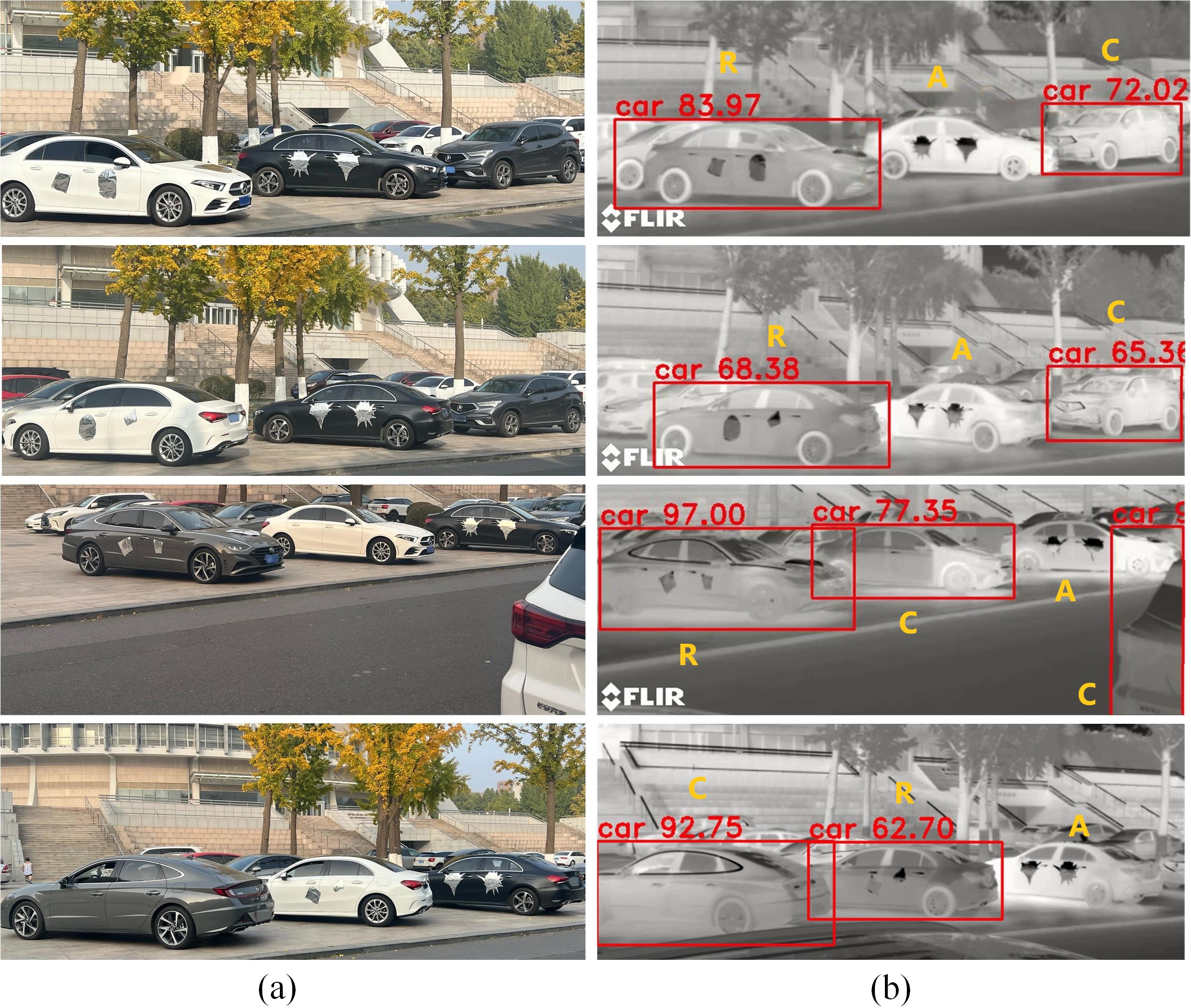} 
\caption{Infrared attack effect on real cars. (a) Visible light view of real cars. (b) Infrared view of real cars. C: clean car. R: car with random shape stickers. A: car with adversarial stickers.
The numbers above the bounding boxes are object confidence scores (\%) with 0.6 threshold. Our adversarial stickers hid the car from Faster RCNN at various viewing angles, 
distances and scenes. In comparison, the clean car and the car with random shape stickers were detected at the same situation. }
\label{one_example}
\end{figure} 

It is well known that deep learning models can be misled by carefully designed perturbations to the input
which is called \textit{adversarial example}, and the perturbation process is called 
\textit{adversarial attack}. Adversarial attacks can be divided into two categories, digital 
attacks \cite{journals/corr/SzegedyZSBEGF13,conf/sp/Carlini017,journals/corr/GoodfellowSS14,conf/iclr/MadryMSTV18,journals/tec/SuVS19,gu2022segpgd} 
which assume that the attackers can directly modify the model input in the digital world,
and physical attacks \cite{conf/cvpr/ThysRG19,zhu2021fooling,hu2022cvpr,wang2022fca,hu2023physically,zhu2022cvpr}
which assume that the attackers can only modify the object or scene in the physical world. 
Physical attacks have attracted much attention because of their importance of assessing 
the security of real-world AI systems.

One type of physical attack is called \textit{infrared physical attack} \cite{zhu2021fooling,zhu2022cvpr,wei2023hotcold,wei2023physically,wei2023unified}.
Infrared imaging is widely used in our daily lives, such as body 
temperature monitoring and autonomous driving.
Since infrared cameras can function normally at night, they are becoming more and 
more important in autonomous driving systems; so is their safety. 

Previous infrared physical attacks \cite{zhu2021fooling,zhu2022cvpr,wei2023hotcold,wei2023physically,wei2023unified} 
mainly focused on infrared pedestrians, and only two works \cite{wei2023physically,wei2023unified} conducted experiments on model cars. 
There is currently a lack of infrared attack research on real cars. The reasons might be as follows. 
Compared with pedestrians with constant body temperature, the temperature distributions of real cars are 
more uneven (e.g., the temperature of the engine is much higher than that of other places), so 
their infrared characteristics are more complex than pedestrians. Compared with model cars without engines, 
real cars' structures and materials are very different from those of model cars, 
so their infrared characteristics are also quite different. Besides, real car attacks require designing 
and manufacturing adversarial patterns on the entire 3D car surface, which poses a great challenge to 
physical experiments. 
But we believe that for the safety of autonomous driving cars, physical real car attack is worth in-depth 
investigation.  

The aforementioned two infrared model car attacks \cite{wei2023physically,wei2023unified} are only effective within limited viewing 
angles (e.g., horizontal angles -$30^{\circ} - 30^{\circ}$ and pitch angle $0^{\circ}$). But we want to 
implement a \textit{full-angle} attack (the horizontal angles $0^{\circ} - 360^{\circ}$, 
and  pitch angle $0^{\circ} - 90^{\circ}$), so the attack angles cover an entire hemisphere surface, 
which is a challenging task. We also notice that these methods \cite{wei2023physically,wei2023unified} 
are both case-by-case attacks, which needs to optimize an adversarial pattern for each 
image\footnote{We found this by checking and running their official codes.}, 
while our goal is to achieve a universal attack which uses the same adversarial pattern to attack detectors at
various viewing angles, distances and scenes.

Towards this goal,  we propose an infrared physical attack method applied to a real car based on 3D modeling. 
We aim to design a set of infrared adversarial stickers to make cars invisible to infrared detectors at various viewing angles (\textit{full-angle}), distances and scenes.
Since most current 3D car models are visible-light models, and there is a lack of 3D infrared car models, 
we build a 3D infrared car model with real infrared characteristics at various viewing angles. 
For the generation of infrared adversarial pattern for stickers, we propose 
to optimize a 3D adversarial mesh at first, then project the shadow of 3D adversarial mesh to obtain a 2D adversarial 
pattern, and finally attach the 2D adversarial pattern to the car surface. 
The motivation of this mesh shadow attack (MSA) method is that we hope to find a better solution in a 
higher-dimensonal 3D space instead of directly optimizing the 2D adversarial pattern. 
To improve the smoothness of adversarial patterns and facilitate sticker implementation, we propose a 3D control 
points-based mesh smoothing algorithm and use a set of smoothing loss functions.
For the physical implementation of infrared adversarial patterns, we use an aluminum film 
which modifies the surface emissivity of the object instead of altering the surface temperature 
used by previous works \cite{zhu2021fooling,zhu2022cvpr,wei2023hotcold,wei2023physically,wei2023unified}.
Like many car stickers, the adversarial car 
stickers can be easily attached on the car surface. The stickers are only 0.08mm thick 
and take up almost no space. 

\begin{figure*}[htbp]
\centering
\includegraphics[scale = 0.52]{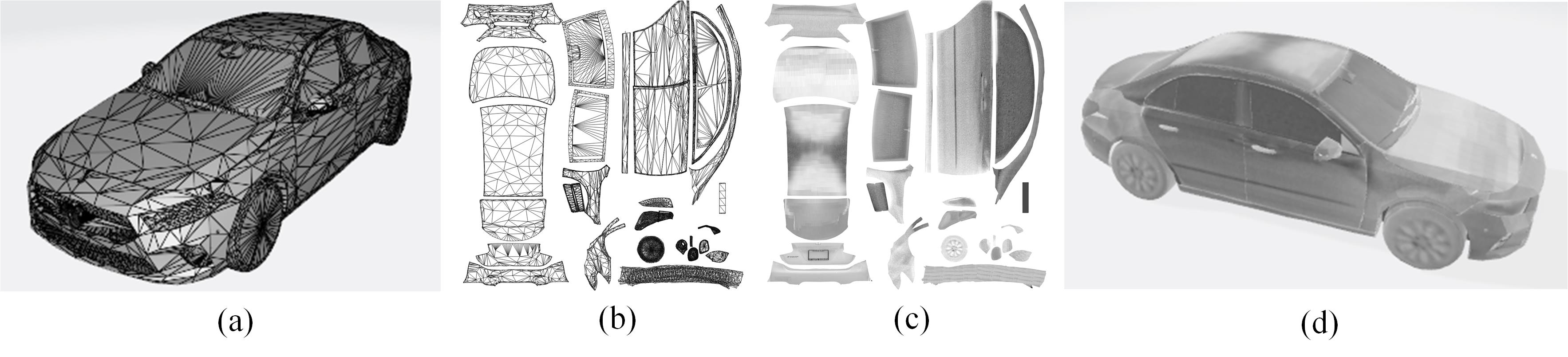} 
\caption{Construction and optimization of real infrared car texture mapping. 
(a) Car mesh model. (b) Reorganized faces map. 
(c) Infrared car texture map collected from real world. (d) Rendered infrared car model.}
\label{car_model}
\end{figure*} 

To assess the safety of infrared 
detection in real autonomous driving scenes, we used two real Mercedes-Benz A200L cars. Physical 
experiments show that our infrared adversarial stickers made the real cars hide from the infrared detector 
Faster RCNN at various viewing angles, various distances, and multiple scenes, with an 
attack success rate (ASR) of 91.49\%. 
To the best our knowledge, this is the first 3D multi-view 
physical infrared vehicle attack, and also the first 
infrared attack conducted on real cars. 

\section{Related Work}
\subsection{Visible Light Physical Adversarial Attacks}
Huang et al. \cite{conf/cvpr/HuangGZXYZL20} proposed a universal physical camouflage (UPC) attack for object detectors. 
Zhang et al. \cite{zhang2019camou} proposed a vehicle camouflage for physical adversarial attack on object detectors in 
the wild. Wang et al. \cite{wang2021dual} proposed the Dual attention suppression (DAS) attack to generate adversarial 
vehicle camouflage in the physical world. Suryanto et al. \cite{suryanto2022dta} generated the physical adversarial 
camouflage by using a differentiable transformation network. Wang et al. \cite{wang2022fca} proposed a 3D full-coverage 
vehicle camouflage for physical adversarial attack (FCA). It is worth noting that all above works are 
proposed for visible light images. 

\subsection{Infrared Physical Adversarial Attacks}
Zhu et al. \cite{zhu2021fooling} proposed a bulb-based board to fool infrared pedestrian detectors in the physical world. 
Zhu at al. \cite{zhu2022cvpr} proposed an infrared invisible clothing to hide from infrared pedestrian detectors
in the physical world. Wei et al. \cite{wei2023hotcold} proposed the HotCold blocks to attack the infrared 
pedestrian detectors. Wei et al. \cite{wei2023physically} proposed a physical adversarial infrared patch (AIP) based on a 
points-clustering algorithm. Wei et al. \cite{wei2023unified} proposed a unified adversarial patch (UAP) for physical attacks 
based on a boundary-limited shape optimization algorithm. It is worth noting that all above methods are 
proposed for infrared pedestrian or model car attacks. There is currently a lack of 
research on infrared real vehicle attacks in the physical world.

\begin{figure}[bp]
\centering
\includegraphics[width=0.8\columnwidth]{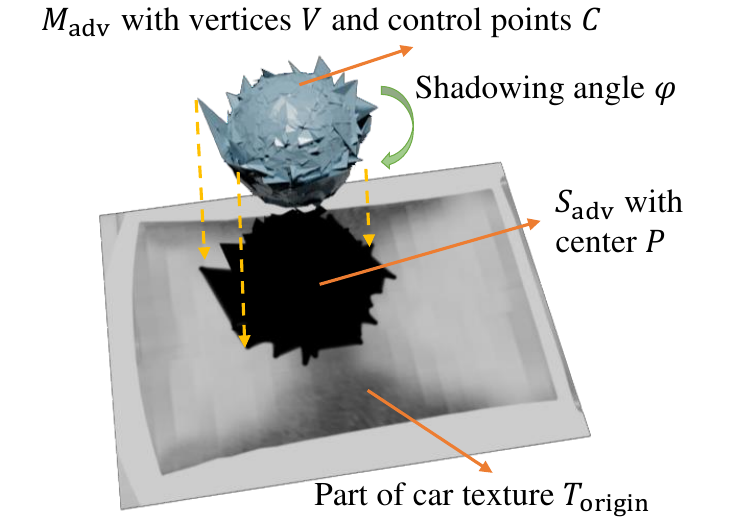} 
\caption{Schematic diagram of mesh shadow method.}
\label{MSA}
\end{figure} 

\begin{figure*}[htbp]
\centering
\includegraphics[scale=0.53]{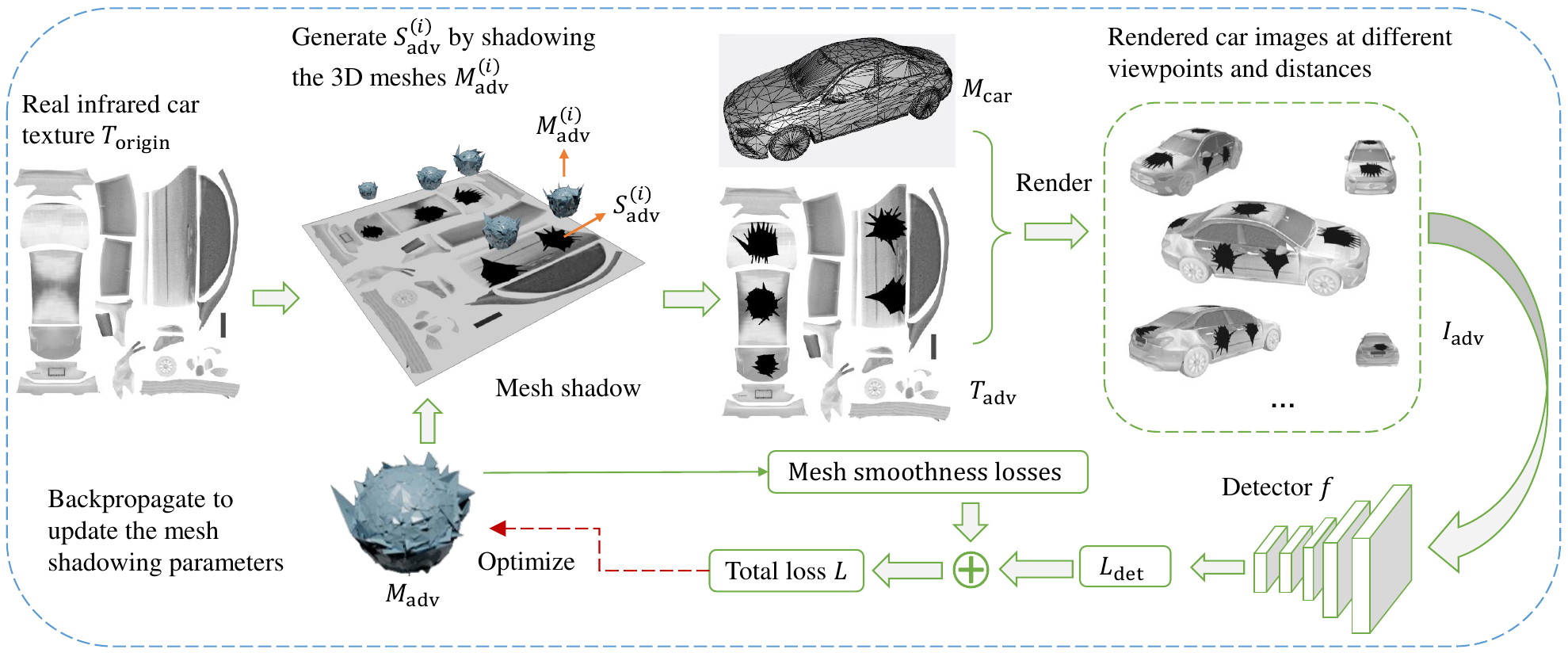} 
\caption{The overall pipeline of the proposed method. }
\label{main_process}
\end{figure*} 

\section{Car Sticker Attack Method} \label{sec:methods}
Our method consist of several steps. First, we build a 3D infrared car model based on real
infrared characteristics. Next, we use the infrared adversarial pattern generation method based on 3D mesh shadow.
To make the 3D adversarial mesh smoother, we use a 3D control points-based mesh smoothing algorithm and use a set of smoothness losses.
Then we apply adversarial patterns to 3D infrared car model and optimize the adversarial patterns.
Finally, we introduce the physical implementation method of infrared adversarial car stickers.

\subsection{Building a 3D Infrared Car Model} \label{sec:car model}
To simulate the infrared car attack realistically, we build a 3D model based on the infrared 
characteristics of a real car. It is worth noting that our method can be applied to any target car, 
and we chose Mercedes-Benz A200L in our experiments, simply because one of the authors have this car.
Figure \ref{car_model}(a) shows the car mesh model ${{M}_{\rm{car}}}$. Next, we need to create 
a ``skin" for the car model based on infrared photos taken by an infrared camera FLIR T560. 
However, the infrared photos captured by the camera are all 2D images, and the challenge is how to ``paste" 
these 2D infrared images onto the 3D car mesh model. First, we flatten all the faces of 3D car mesh 
onto a 2D plane called \textit{faces map}. After that, we use MAYA software to rearrange these faces to divide 
different areas, such as roof, doors, etc., as shown in Figure \ref{car_model}(b). This process facilitates the 
alignment of real infrared car images with the 3D car mesh. 

Subsequently, we crop the infrared 
images into different parts based on the \textit{faces map} (Figure \ref{car_model}(b)) and paste the cropped images onto the 
2D \textit{faces map}, and then we get the infrared \textit{texture map} of the car, as shown in Figure \ref{car_model}(c). 
See \textit{Supplementary Material (SM)} for how these infrared photos are taken, cropped and pasted.
This process establishes a correspondence between the real infrared car images and the 3D car 
mesh.  The rendered infrared car model is shown in Figure \ref{car_model}(d), which is built for a real car with engine running. 


\subsection{Generation of 2D Shadow Based on 3D Mesh}
We aim to design infrared stickers with adversarial patterns to hide the cars from infrared detectors 
at various viewing angles (\textit{full-angle}), distances, and scenes.
We propose a 3D adversarial mesh shadow 
attack (MSA) method to generate the 2D infrared adversarial patterns for stickers. The motivation of MSA method is that 
we hope to find a better solution in a higher-dimensonal 3D space instead of directly optimizing the 
2D adversarial pattern. 
The core of MSA method is to optimize 
a 3D adversarial mesh ${{M}_{\rm{adv}}}$ at first, then project the shadow of 3D adversarial mesh to obtain a 2D adversarial pattern ${{S}_{\rm{adv}}}$, 
and finally attach the 2D adversarial pattern ${{S}_{\rm{adv}}}$ to the car surface. Note that the 3D adversarial mesh ${{M}_{\rm{adv}}}$ is different 
from the 3D car mesh ${{M}_{\rm{car}}}$ in Section \ref{sec:car model}. The optimization variables include 
the 3D mesh vertices coordinates $V$, the mesh shadowing angle $\varphi$, and the center point 
position $P$ when pasting the 2D pattern onto the car's texture map ${T}_{\rm{origin}}$ (Figure \ref{MSA}). 

The shadowing operation refers to rendering the mesh ${{M}_{\rm{adv}}}$ to a dark area while 
retaining the mesh contour (Figure \ref{MSA}), and the dark area has a uniform 
grayscale value within the contour. The grayscale value is consistent with the infrared 
characteristics of the sticker we use. 
Let $\Omega$ denote this operation. If we want the car to  
have adversarial effect at various viewing angles, we need to optimize 
$N$ adversarial meshes to generate $N$ adversarial shadows on different places of the car (Figure \ref{main_process}):
\begin{equation}
	S_{\rm{adv}}^{\left( i \right)}=\Omega(M_{\rm{adv}}^{\left( i \right)},{{\varphi }^{\left( i \right)}}),i=1,2,...N.
\end{equation}


\subsection{3D Control Points-Based Mesh Smoothing}
If we directly optimize the vertices coordinates $V$ of the mesh ${{M}_{\rm{adv}}}$, many ``peaks" will appear 
on the mesh surface, which will make the shadow shape very complex and will be difficult for the physical 
implementation of the adversarial shadow ${{S}_{\rm{adv}}}$. Inspired by the Gaussian smoothing kernel and 
spline interpolation method, we propose a smoothing control algorithm for 3D mesh vertices. Its core 
idea is to use some 3D control points $C$ as anchor points, and the 
coordinate offsets of mesh vertices $V$ are expressed as the weighted average of the coordinate offsets of $C$.
The calculation details are described in \textit{SM}. 

We use $\Theta$ to denote the above transformation, and $C^{\left( i \right)}$ to denote 
the control points set of $M_{\rm{adv}}^{\left( i \right)}$, then
\begin{equation}
	{V^{\left( i \right)}}=\Theta(C^{\left( i \right)}), i=1,2,...N.
\end{equation}
Since the control points $C^{\left( i \right)}$ determine the vertices coordinates $V^{\left( i \right)}$, the optimization variables changes 
from $\left( V^{\left( i \right)},P^{\left( i \right)},\varphi^{\left( i \right)} \right)$ to $\left( C^{\left( i \right)},P^{\left( i \right)},\varphi^{\left( i \right)} \right), i=1,2,...N$. 

\subsection{Mesh Smoothness Loss Functions}
To further enhance the smoothness of 3D adversarial mesh ${{M}_{\rm{adv}}}$ and 2D adversarial pattern 
${{S}_{\rm{adv}}}$, we use a set of loss functions including: \textit{mesh normal consistency loss}, \textit{mesh edge loss}, 
\textit{chamfer distance loss}, and \textit{Laplacian smoothing loss}.
During the optimization process of 3D mesh ${{M}_{\rm{adv}}}$ , these functions guide the generation of 
a smoother adversarial mesh. A smoother 3D mesh ${{M}_{\rm{adv}}}$ results in a smoother 2D shadow pattern 
${{S}_{\rm{adv}}}$, which is beneficial for manufacturing physical stickers based on the 2D shadow 
pattern. 

The \textit{mesh normal consistency loss} computes the normal consistency for each pair of neighboring faces, and minimizing it
encourages the mesh surface to be smoother. We suppose that mesh ${{M}_{\rm{adv}}}$ has a total of $F$ faces.
Let ${{n}_{i}},{{n}_{j}}$ ($1\le i,j\le F $) represent the normal vector of any two adjacent faces, then this 
loss function is described as:
\begin{equation}
	{{L}_{\rm{norm}}}=\mathrm{Average}\left( 1-\cos ({{n}_{i}},{{n}_{j}}) \right).
\end{equation}
$\mathrm{Average}$ is calculated between any pair of adjacent faces. 

The \textit{mesh edge loss} computes mesh edge length regularization loss, and minimizing it encourages a reduction in the average edge 
length of the adversarial mesh. Suppose ${{M}_{\rm{adv}}}$ has a total of $M$ edges, ${{l}_{k}}$ ($k=1,2,...,M$) represents the 
length of each edge, and this loss function is described as
\begin{equation}
    {{L}_{\rm{edge}}}={\left( \sum\limits_{k=1}^{M}{{{l}_{k}}} \right)}/{M}.
\end{equation}

The \textit{chamfer distance loss} computes chamfer distance \cite{fan2017point} between the sampling points of adversarial mesh and a standard 
sphere. Reducing chamfer distance loss encourages the adversarial mesh ${{M}_{\rm{adv}}}$ to approximate a 
sphere. We denote the point clouds sampled by adversarial mesh ${{M}_{\rm{adv}}}$ and the standard spherical 
mesh ${{M}_{\rm{sphere}}}$ by ${{S}_{1}}$ and ${{S}_{2}}$, respectively. The chamfer distance loss is described 
as 
\begin{equation}
	{{L}_{\rm{chamfer}}}=\sum\limits_{p\in {{S}_{1}}}{\underset{q\in {{S}_{2}}}{\mathop{\min }}\,}\left\| p-q \right\|_{2}^{2}+\sum\limits_{q\in {{S}_{2}}}{\underset{p\in {{S}_{1}}}{\mathop{\min }}\,}\left\| q-p \right\|_{2}^{2}.
\end{equation}

The \textit{Laplacian smoothing loss} ${{L}_{\rm{Laplace}}}$ computes the Laplacian smoothing objective for the adversarial mesh. 
We define this function 
as $\mathrm{Laplace}$, and its calculation details are introduced in \cite{nealen2006laplacian}, so
\begin{equation}
	{{L}_{\rm{Laplace}}}=\mathrm{Laplace}\left( {{M}_{\rm{adv}}} \right).
\end{equation}

\subsection{Applying the 2D Shadow to 3D Car Model}

We apply the adversarial shadow ${{S}_{\rm{adv}}}$ to 3D infrared car model by changing its texture map ${{T}_{\rm{origin}}}$. 
This simulates the process we paste the adversarial stickers to the car surface in the physical world. To facilitate physical 
implementation, we simulate to paste the stickers onto the door, roof, front hood and rear of the car, 
which have wide ranges of flat area and are easy to paste. In each area, we paste one or two adversarial shadow 
patterns (Figure \ref{main_process}). To simulate real-world perturbations, such as errors in cutting the adversarial stickers, 
variations in surface temperature on the adversarial stickers, and errors in the pasting positions, we 
introduce random perturbations to the vertex coordinates $V$ of the adversarial mesh ${{M}_{\rm{adv}}}$, 
random noise to the pattern ${{S}_{\rm{adv}}}$, random changes in grayscale values of ${{S}_{\rm{adv}}}$, and 
random perturbations in the position $P$ during the pasting of ${{S}_{\rm{adv}}}$. This approach, known as the 
Expectation Over Transformation (EOT) algorithm \cite{conf/icml/AthalyeEIK18}, enhances the robustness of our algorithm in 
real-world scenes.

Next, we need to paste the adversarial patterns ${{S}_{\rm{adv}}}$ onto the original car texture map ${{T}_{\rm{origin}}}$. 
Let $\Gamma$ denote the pasting operation.
We establish a Cartesian coordinate system with the center point of ${{T}_{\rm{origin}}}$ as 
the origin. We use  ${P}^{\left( i \right)}$ to 
represent the coordinates of the pasting positions for $N$ adversarial shadows
$S_{\rm{adv}}^{\left( i \right)}, i=1,2,...N$. The texture map after pasting the adversarial shadows is 
\begin{equation} \label{equ10}
	{{T}_{\rm{adv}}}=\Gamma\left( S_{\rm{adv}}^{\left( i \right)},{{P}^{\left( i \right)}},{{T}_{\rm{origin}}} \right),i=1,2,...N.
\end{equation}

We use the differentiable renderer Pytorch3D \cite{ravi2020pytorch3d} to render the adversarial texture map ${{T}_{\rm{adv}}}$ onto the surface of 
the car mesh ${{M}_{\rm{car}}}$, resulting in the rendered infrared car images with adversarial patterns, 
denoted as ${{I}_{\rm{adv}}}$. Let $\Psi$ denote the rendering operation with parameters $\theta$ which include the rendering distances and angles. 
Mathematically, this process can be 
expressed as:
\begin{equation}
	{{I}_{\rm{adv}}}=\Psi\left( {{M}_{\rm{car}}},{{T}_{\rm{adv}}},\theta  \right).
\end{equation}

\subsection{Optimization of 3D Mesh Shadow Attack}
After we get the rendered infrared adversarial images ${{I}_{\rm{adv}}}$, we input them into the target 
detector $f$. The output of the target detector typically includes object confidence ${f}_{\rm{obj}}$, class 
confidence ${f}_{\rm{cls}}$, and bounding box ${f}_{\rm{bbox}}$. Since our goal is to create a stealthy attack, 
meaning that our adversarial texture ${{T}_{\rm{adv}}}$ should make the infrared car hide from the detector, 
we try to lower the object confidence ${f}_{\rm{obj}}\left( {{I}_{\rm{adv}}} \right)$ as much as possible. 
Therefore, the detection loss is defined as:
\begin{equation}
	{{L}_{\rm{det}}}={{f}_{\rm{obj}}}\left( {{I}_{\rm{adv}}} \right).
\end{equation}

The overall loss function is defined as follows:
\begin{equation} \label{equ_loss}
\begin{split}
    L ={{L}_{\rm{det}}}+{{w}_{1}}\cdot {{L}_{\rm{norm}}}+{{w}_{2}}\cdot {{L}_{\rm{edge}}} \\ 
    +{{w}_{3}}\cdot {{L}_{\rm{chamfer}}}+{{w}_{4}}\cdot {{L}_{\rm{Laplace}}}.
\end{split}
\end{equation}
Here, $w_{1}$, $w_{2}$, $w_{3}$, and $w_{4}$ are weights of different losses, which are determined empirically. 
We use the backpropagation algorithm according to the loss function $L$ to update the optimization 
variables ${{C}^{\left( i \right)}},{{P}^{\left( i \right)}},{{\varphi }^{\left( i \right)}}$, $i=1,2,…,N$.
The overall pipeline is illustrated in Figure \ref{main_process}.

\subsection{Physical Implementation Method}

We use aluminum films to make adversarial car stickers. 
Instead of altering the surface temperature of an object, this approach focuses on modifying the surface emissivity of the object, 
which is different from previous works \cite{zhu2021fooling,zhu2022cvpr,wei2023hotcold,wei2023physically,wei2023unified}. 
Aluminum typically has an emissivity around 0.1, while the surface of a car, 
typically made of steel, has an emissivity around 0.8, resulting in different infrared characteristics. 
We utilize an ultra-thin (only 0.08mm) aluminum film, which can be easily attached on the 
surface of a car like many car stickers.  
We only need around 13 mins to make a sticker, and the cost of a sticker is only around 0.2 USD.
The implementation process of adversarial stickers is shown in \textit{Supplementary Video 1}. 


\begin{table*}[htbp]
\centering
\caption{ASRs (\%) of cars with different textures against different detectors. }\label{tab_asr}
\begin{tabular}{c|ccccccc}
\toprule[1.1pt]  
\diagbox{Texture}{Detector} & Faster & RetinaNet & Cascade  & Libra & SSD & YOLOv3 & Deformable \\
\midrule  
Origin& 2.10 & 4.49 & 18.86 & 16.47 & 15.27 & 25.15 & 12.28 \\
Random shape& 18.26 & 17.07 & 23.95 & 28.74 & 45.81 & 50.00 & 23.65 \\
AIP & 14.97 & 15.27 & 32.93 & 27.54 & 23.65 & 38.32 & 27.54 \\
UAP & 20.06 & 31.44 & 38.02 & 33.23 & 36.83 & 39.82 & 24.25 \\
Ours& \textbf{96.31} & \textbf{86.83} & \textbf{73.35} & \textbf{79.04} & \textbf{95.80} & \textbf{86.52} & \textbf{83.83} \\
\bottomrule[1.1pt] 
\end{tabular}

\end{table*}

\begin{figure}[bp]
\centering
\includegraphics[width=1\columnwidth]{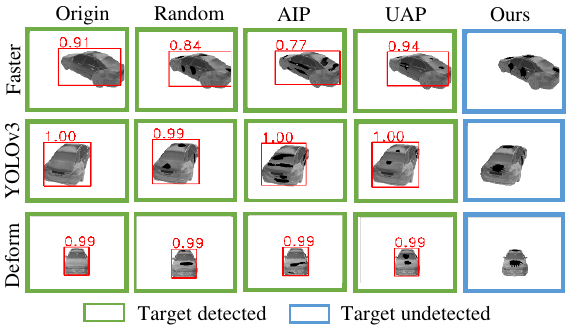} 
\caption{Examples of detection results of different detectors for target cars with different textures. 
The numbers above the red bounding boxes are the object confidence scores, with a threshold of 0.6.
The results of other detectors are shown in \textit{SM}. }
\label{many_exps}
\end{figure} 

\begin{figure*}[htbp]
\centering
\includegraphics[scale = 0.66]{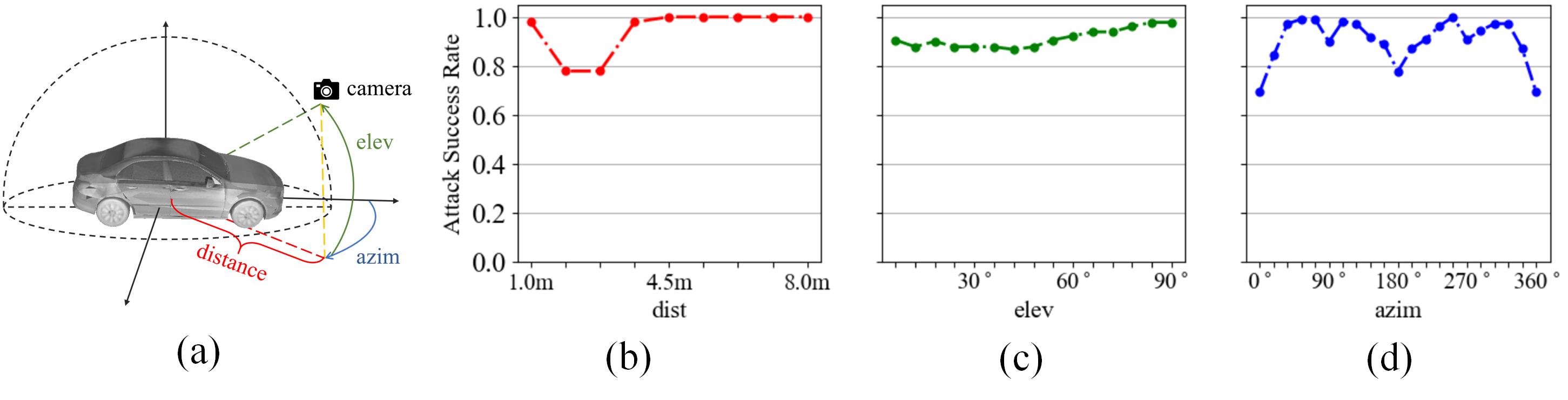} 
\caption{ Full-angle ASRs at diffrent (a) parameters including (b) distances, (c) pitch angles, and (d) horizontal angles. See text for details.  }
\label{angle-dist}
\end{figure*} 

\section{Experiments}
\subsection{Dataset}
We used the FLIR\_ADAS\_1\_3 \cite{FLIR_ADAS} infrared dataset released by FLIR company for infrared autonomous driving 
scenes. It contains 10,228 real infrared photos collected in streets and highways of Santa Barbara, 
USA. The infrared camera is FLIR Tau2. The training set 
contains 7160 images, and the test set contains 3068 images. 
We used this dataset to finetune the target detector.

\subsection{Target Detectors}
We initially chose the classic two-stage object detector: Faster R-CNN \cite{ren2016faster}, as our primary target detector. 
We used the pre-trained Faster RCNN model provided by the torchvison library \cite{torchvision2016} and then finetuned it 
on the FLIR\_ADAS\_1\_3 dataset. The average precision (AP) for car class of the finetuned model on the training set was 0.96, and the AP on 
the test set was 0.92. After attacking Faster RCNN in a white-box setting, we transfered our attack method 
to other unseen detectors such as YOLOv3 \cite{journals/corr/abs-1804-02767}, Deformable DETR \cite{conf/iclr/ZhuSLLWD21}, etc. 
which were provided by mmdetection library \cite{mmdetection} in a black-box setting. 

\subsection{Evaluation Metrics}
In our experiments, we used the attack success rate (ASR) as the evaluation metrics of the attack methods.
The ASR was defined as the ratio of the number of cars which were not detected to the total number of cars.
We set the confidence threshold of target detectors to 0.6 and the IOU threshold between the prediction box and 
ground truth to 0.5, similar to previous works \cite{zhu2021fooling,zhu2022cvpr,wei2023physically,wei2023unified}.
The ASR was calculated based on the average value of sample points collected from various distances, horizontal angles, and pitch angles
with the sampling method described in Section \ref{sec:faster}.

\subsection{Attack Faster RCNN in the Digital World} \label{sec:faster}

We optimized $N=5$ adversarial shadow patterns to simulate 
5 adversarial stickers to be pasted on the car surface. 
The hyper-parameters are detailed in \textit{SM}.  
After optimization, we obtained the adversarial shadow patterns (${{T}_{\rm{adv}}}$ in Figure \ref{main_process}), 
and the rendered car with adversarial shadow patterns (${{I}_{\rm{adv}}}$ in Figure \ref{main_process}).

After that, we evaluated the attack effectiveness of the adversarial shadow patterns. 
For a fair comparison, 
we employed the original car pattern (without any sticker) and random shape pattern
as control patterns. The figures of these patterns are shown in \textit{SM}. These patterns were 
rendered onto the same car model, and the resulting images 
were input into Faster R-CNN.  
The results, as shown in Table \ref{tab_asr}, indicate that our adversarial 
shadow patterns achieved an ASR of 96.31\% for Faster R-CNN in the digital world. In comparison, the ASRs for the 
original car pattern and random shape pattern were only 2.10\% and 18.26\%, respectively. This demonstrates the 
effectiveness of our attack method. Figure \ref{many_exps} shows typical examples.

We then analyze the ASR of our method at various (\textit{full-angle}) viewing angles and distances. 
The horizontal angle $azim$ ranged from 0 to 360 degrees, and we sampled it every 20 degrees. The pitch 
angle $elev$ ranged from 0 to 90 degrees, and we sampled it every 6 degrees. The distance $dist$ ranged from 
1 to 8 meters, and we sampled it every 1 meter. 
Figure \ref{angle-dist}(b-d) show the ASRs with respect to one variable (e.g., $dist$) by taking the 
average of ASRs over all combinations of values of the other two variables (e.g., $elev$ and $azim$). 
The results indicate that our approach achieves successful 
attacks at various (\textit{full-angle}) viewing angles and various distances. In contrast, many
previous works \cite{zhu2021fooling,wei2023hotcold,wei2023physically,wei2023unified} were effective only within limited viewing angles (e.g., horizontal angle from 
-30 to 30 degrees and pitch angle 0 degree) and shorter distances (e.g. 3 to 6 meters). 
Note that there is a decrease in ASR at 2 meters and 0(or 180)-degree horizontal angle. This suggests 
that Faster RCNN is more robust in these specific scenes, potentially due to the distribution of 
training images. Nevertheless, for the majority of viewing angles and distances, the ASRs of our method consistently 
exceeded 80\%.


\subsection{Ablation Study}

To evaluate the effectiveness of the 3D control points-based mesh smoothing algorithm (CMS) and a set of 
smoothing losses (SMLS), we performed ablation experiments. We conducted a subjective evaluation on the 
smoothness score of the 3D adversarial mesh and 2D adversarial pattern and  we also evaluated the 
physical implementation time of 2D adversarial patterns under different settings. The experimental settings
and results are detailed in \textit{SM}. The results indicate that both CMS and SMLS improved the smoothness of adversarial meshes and patterns, 
and their combination was better. Besides, these methods effectively reduced the physical implementation 
time of adversarial patterns.

\subsection{Exploring the Interpretability of the Attack}
To gain deeper insights into our attack methods, we utilized the GradCAM \cite{selvaraju2017grad} technique to analyze the 
changes in network attention maps before and after the attack. See \textit{SM} for more details.


\begin{figure}[bp]
\centering
\includegraphics[width=1\columnwidth]{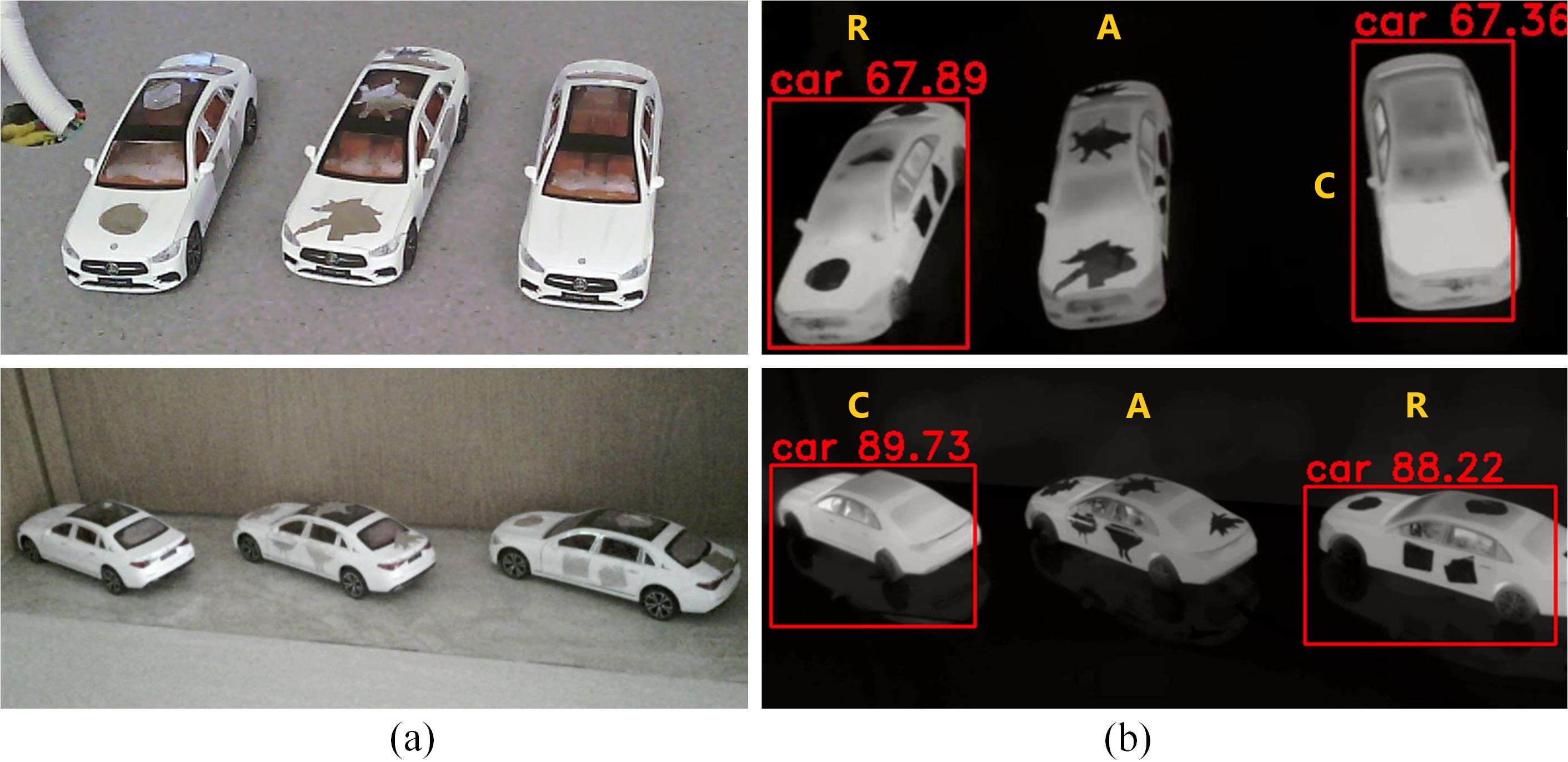} 
\caption{Infrared attack effect on model cars. (a) Visible light view of model cars. 
(b) Infrared view of model cars. C: clean car. R: car with random shape stickers. 
A: car with adversarial stickers.}
\label{model_car}
\end{figure} 

\begin{figure*}[htbp]
\centering
\includegraphics[scale=0.153]{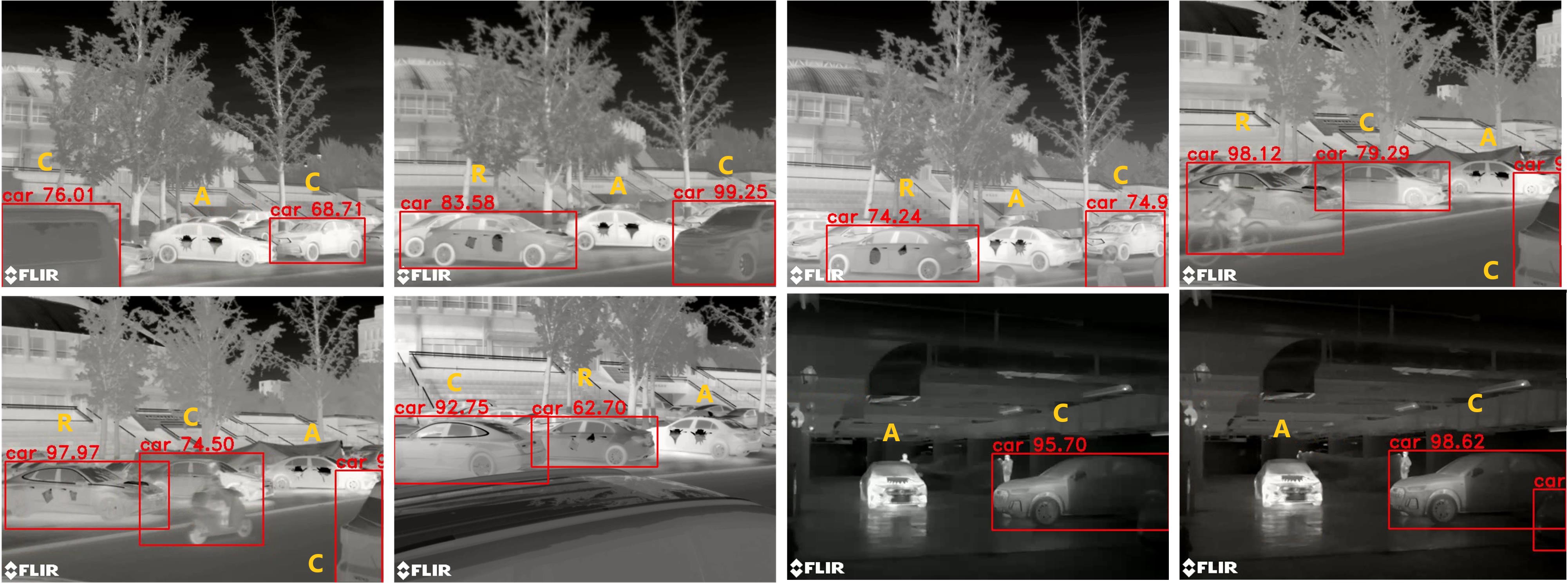} 
\caption{ Examples of infrared real car attacks. C: clean car. R: car with random shape stickers. A: car with adversarial stickers. }
\label{phy_exps}
\end{figure*} 



\subsection{Comparison with 2D Optimization Methods}
We extended the previous 2D infrared model car attack methods \cite{wei2023physically,wei2023unified} 
to our 3D car model.
We generated adversarial car textures on our car model based on their papers and codes, which are shown in 
\textit{SM}. Following the settings in Section \ref{sec:faster}, we evaluated the attack performance of the different 
methods. The statistics of ASRs are presented in Table \ref{tab_asr}, with typical examples shown in Figure \ref{many_exps}.


The results indicated that the ASR of our method (96.31\%) outperformed the ASRs of two alternative 
methods (14.97\% and 20.06\%) for Faster RCNN in the digital world. The reasons may be as follows.
The two 2D optimization methods based on boundary optimization \cite{wei2023unified} or points 
clustering \cite{wei2023physically} are subject to various constraints. 
These constraints are used to ensure that, for example, the boundary curves do not have overlaps \cite{wei2023unified}, 
and the adversarial patterns do not split into multiple pieces \cite{wei2023physically}. 
With more constraints, the feasible solution space becomes smaller.  
However, our 3D mesh shadow approach does not have such constraints, and we can
explore a larger solution space, leading to better results.

\subsection{Attack Transferability}
We tested the effectiveness of our adversarial car texture (${{T}_{\rm{adv}}}$ in Figure \ref{main_process}) optimized for Faster RCNN 
against other unseen detectors, including RetinaNet \cite{conf/iccv/LinGGHD17}, Cascade RCNN \cite{cai2019cascade}, 
Libra RCNN \cite{pang2019libra}, SSD \cite{liu2016ssd}, YOLOv3 \cite{journals/corr/abs-1804-02767} and Deformable DETR \cite{conf/iclr/ZhuSLLWD21}. 
It is worth noting that these experiments were 
performed in a black-box setting, which is a more challenging but practically significant scene for 
real-world applications. The results are shown in Table \ref{tab_asr}. The ASRs of our method against unseen models 
reached 73.35\%-95.80\%. It indicates that our method performed well in a black-box setting and had good attack
transferability to unseen models including not only CNN-based models but also transformer-based models. 
Besides, the transferability of our method is stronger than not only two simple baselines but also 
two infrared attack methods \cite{wei2023physically,wei2023unified}.

\subsection{Physical Attacks on Model Cars} \label{sec:model}
We initially conducted physical experiments on three same 1:24 scale Mercedes-Benz model cars 
(Figure \ref{model_car}(a)). We crafted the 
adversarial aluminum stickers according to the optimized patterns, scaled to match the size of the model 
car, and applied them to the model car. 
In addition, we created randomly cut-shaped stickers as a control.
We heated the model cars with hot water to simulate the real car with engine running.
We used a rotating turntable to conveniently 
assess the attack effectiveness over the entire 0-360 horizontal degree and 0-90 pitch angle range. 
The infrared camera was FLIR T560. We utilized the same Faster RCNN detector 
as in Section \ref{sec:faster}. The detection results indicated that our adversarial 
stickers achieved an ASR of 84.86\% on the 1:24 scale car model in the physical world.  In comparison, the 
clean car and the car with random patterns had ASRs of only 19.08\% and 35.37\%, respectively. Figure \ref{model_car} 
provides specific examples from the physical world experiments. A demo video for physical model car attacks
is shown in \textit{Supplementary Video 2}.

\subsection{Physical Attacks on Real Cars} \label{sec:real}

We conducted physical experiments on two real Mercedes-Benz A200L cars (one black one
white). It was a sunny day with a temperature of about 25$^{\circ}$C. For a fair comparison, 
we pasted the the adversarial stickers, randomly cut-shaped stickers or nothing on the same car successively.
We recorded 30 videos (each video is 
around 2 minutes) and sampled 3688 infrared images of these cars from various angles and distances in 
both ground and underground parking lots using a FLIR T560 camera. The height of the camera tripod can be 
adjusted from 1m to 2m. We sent these images to Faster RCNN. 

The results indicated that our adversarial stickers achieved an ASR of 91.49\% on the real 
cars with the engines running, while the random shape stickers and no sticker had ASRs of only 6.21\% 
and 0.66\%, respectively. When the engines were off for one hour, the ASR of adversarial stickers dropped a little 
to 88.42\%. The reason might be that the infrared patterns of adversarial stickers with engines off were not as clear 
as the patterns with engines running.
However, the ASR of adversarial stickers were still better than ASR of random shape 
stickers (5.72\%) and no sticker (1.86\%) when the engines were off. Figure \ref{one_example} and 
Figure \ref{phy_exps} show some examples. 
There were a few other cars that passed by or were parked when we were recording videos and therefore 
appeared in our photos, which were also detected.
See \textit{Supplementary Video 3} for the demo video.


\subsection{Adversarial Defense}
We tested five adversarial defense methods to defend our attack methods in the digital world, including adversarial training \cite{journals/corr/GoodfellowSS14}, PixelMask \cite{agarwal2021cognitive}, 
Bit squeezing \cite{conf/ndss/Xu0Q18}, JPEG compression \cite{conf/iclr/GuoRCM18} and Total variation minimization \cite{conf/iclr/GuoRCM18}. Experiment 
details for these methods are in \textit{SM}. The results show that although these methods had a certain defense 
effect, the ASR of our method still reached 88.83\%-94.81\% after adding defense, which shows 
that our method is a powerful attack method.


\section{Conclusion}
We propose infrared adversarial stickers to hide a real car from 
infrared detectors at various viewing angles, distances, and scenes in the physical world. 
We build a 3D infrared car model with real infrared characteristics and propose a 3D mesh shadow method
for the generation of infrared adversarial pattern.
To make the 3D adversarial mesh smoother, we propose a 
3D control points-based smoothing algorithm and use a set of smoothness loss functions.
Our adversarial stickers enabled two real cars to evade Faster RCNN at various viewing angles, distances and scenes.
Besides, our method has strong attack transferability against multiple unseen detectors in a black-box setting.


\section{Acknowledgements}
This work was supported in part by the National Natural Science Foundation of 
China (Nos. 61734004, U2341228, U19B2034).


{
    \small
    \bibliographystyle{ieeenat_fullname}
    \bibliography{egbib}
}


\end{document}